\pdfoutput=1

\documentclass[11pt]{article}

\usepackage{ACL2023}

\usepackage{times}
\usepackage{latexsym}
\usepackage[greek, english]{babel}
\usepackage{booktabs}
\usepackage{tabularx}
\usepackage{multirow}
\renewcommand{\arraystretch}{0.9}
\usepackage{relsize}
\usepackage{enumitem}
\setlist[itemize]{leftmargin=*}
\setlist[enumerate]{leftmargin=*}
\setlist{noitemsep}
\usepackage{amssymb}

\usepackage[T1]{fontenc}
\usepackage[utf8]{inputenc}
\usepackage{microtype}
\usepackage{inconsolata}

\title{OYXOY: A Modern NLP Test Suite for Modern Greek}

\newcommand{\afa}{\textsuperscript{\textnormal 1}\,}
\newcommand{\afb}{\textsuperscript{\textnormal 2}\,}

\author{
    \AND
    Konstantinos Kogkalidis\afa$^\star$ \\ {\texttt{kokos.kogkalidis@aalto.fi}} \And Stergios Chatzikyriakidis\afb$^\star$ \\ {\texttt{stergios.chatzikyriakidis@uoc.gr}}
    \AND
    Eirini Chrysovalantou Giannikouri\afb \And Vassiliki Katsouli\afb \And Christina Klironomou\afb 
    \AND Christina Koula\afb  \And  Dimitris Papadakis \afb  \And Thelka Pasparaki\afb
    \AND Erofili Psaltaki\afb \And Efthymia Sakellariou\afb \And Hara Soupiona\afb
    \\
    \AND
    \\
    \afa Department of Computer Science, Aalto University\\
    \afb Department of Philology, University of Crete\\
    \addlinespace
    $^\star$ Corresponding
}

\newcommand{\num}[2]{\ensuremath{#1{\mathsmaller{\pm#2\%}}}}

\begin{document}

\maketitle
\begin{abstract}
This paper serves as a foundational step towards the development of a linguistically motivated and technically relevant evaluation suite for Greek NLP.
We initiate this endeavor by introducing four expert-verified evaluation tasks, specifically targeted at natural language inference, word sense disambiguation (through example comparison or sense selection) and metaphor detection.
More than language-adapted replicas of existing tasks, we contribute two innovations which will resonate with the broader resource and evaluation community.
Firstly, our inference dataset is the first of its kind, marking not just \textit{one}, but rather \textit{all} possible inference labels, accounting for possible shifts due to e.g. ambiguity or polysemy.
Secondly, we demonstrate a cost-efficient method to obtain datasets for under-resourced languages. 
Using ChatGPT as a language-neutral parser, we transform the Dictionary of Standard Modern Greek into a structured format, from which we derive the other three tasks through simple projections.
Alongside each task, we conduct experiments using  currently available state of the art machinery.
Our experimental baselines affirm the challenging nature of our tasks and highlight the need for expedited progress in order for the Greek NLP ecosystem to keep pace with contemporary mainstream research.
\end{abstract}

\section{Introduction}
It is a well known fact that the natural language processing world is running at multiple speeds.
A select few languages claim the lion's share in the literature, boasting a plethora of models and a constant stream of results, while others are struggling to keep up with last year's state of the art.
Meanwhile, multilingual models, despite being heralded as the end-all solution to the issue, often fall short of expectations~\cite[\textit{inter alia}]{wu-dredze-2020-languages,ogueji-etal-2021-small,pfeiffer-etal-2021-unks,espana-bonet-barron-cedeno-2022-undesired,havaldar-etal-2023-multilingual,papadimitriou-etal-2023-multilingual-bert}.
The assumption that one-size-fits-all multilingual models can effectively bridge the language gap is hard to either refute or validate, given the disproportionate distribution of training and evaluation resources among languages~\cite{joshi-etal-2020-state,yu-etal-2022-beyond,10.1162/tacl_a_00447}.
Further muddying the waters is the dubious quality of the increasingly trending multi- and mono-lingual resources generated through minimally supervised machine translations from English~\cite{artetxe-etal-2020-translation,wang-hershcovich-2023-evaluating}.
While such endeavors can certainly make for good first steps, they are neither sufficient nor without risks.
The wide adoption of the practice threatens resource plurality, as more and more ``new'' datasets are in fact old in all but language.
Furthermore, it condones the accumulation of academic authority to a select few, namely the authors of the originals, promoting the unhindered perpetuation of their biases and oversights as universal across languages.
Worse yet, it outsources linguistic expertise to machine labor, as we are now entrusting our automated processes with capturing the nuances of under-represented languages; exactly \textit{those} languages that require opinionated and targeted expert attention the most.

And while a discussion on the structural causes behind the problem and the ways to incentivize change is long overdue, here we set our aims towards something more actionable.
Noting the striking absence of evaluation benchmarks for modern Greek, and the language's limited presence in multi-lingual resources, we set out to develop a linguistically motivated and technically relevant suite of evaluation tasks.
This paper aims to kickstart this endeavor, while serving as an open invitation to interested parties.
Concretely, we set the pace with four evaluation tasks:
\begin{enumerate}[noitemsep,topsep=0pt]
    \item a handcrafted dataset for inference, consisting of 1\,762 sentence pairs, each pair adorned with a linguistic characterization in the form of tags \textit{\`a la} SuperGlue and labeled with a subset (rather than an element) of \{\texttt{Neutral}, \texttt{Entailment}, \texttt{Contradiction}\}, aiming to account for all possible inference relations between premise and hypothesis
    \item a structured translation of the Dictionary of Standard Modern Greek, from which we project into three tasks:
    \begin{enumerate}[noitemsep, topsep=0pt]
        \item[(i)] a word sense disambiguation task \textit{\`a la} Words-in-Context, consisting of 117\,662 phrase pairs that correspond to two usage examples for a single word, where the system is tasked with telling whether the two occurrences have the same meaning or not
        \item[(ii)] a more compact \& linguistically informed version of the same task consisting of 14\,416 unique phrases containing polysemous words, each word associated to a number of senses and their periphrastic definitions, where the system is tasked with telling which word sense is associated with each usage example
        \item[(iii)] a metaphor detection task, associating each of the previous phrases to a boolean label indicating whether the word in focus is used metaphorically or not
    \end{enumerate}
\end{enumerate}
To facilitate research with these tasks, we supply accessible entry points to the raw data in the form of Python interfaces.
For each task, we conduct experiments using the currently available state of the art machinery and establish baseline scores for comparisons.%
\footnote{
Data, interfaces and the code necessary to replicate our experiments is available at \href{https://github.com/StergiosCha/OYXOY}{github.com/StergiosCha/OYXOY}.
}

\section{OYXOY}
Inspired by Glue and SuperGlue~\cite{GLUE,SGLUE}, our goal is to develop a language-adapted suite that selects and extends a few key aspects of the original. 
Our project, which we lightly dub OYXOY (pronounced /\'uxu/), is not primarily focused on offering general diagnostics, but rather on highlighting the semantic, syntactic, and morphological attributes of the Greek language, and quantifying their impact on NLP systems.
To that end, we present four high-level tasks that require varying degrees of lexical \& sentential meaning comprehension.

\subsection{Natural Language Inference}
Our first task is a staple of computational semantics that has endured the test of time: natural language inference (NLI).
In their most common form, NLI tasks present the system with an ordered pair of sentences (called a premise and a hypothesis), and request one of three inference relations that must hold between premise to hypothesis: \texttt{Entailment}, \texttt{Contradiction} and \texttt{Neutral}/\texttt{Unknown}. 
Despite its apparent simplicity and the heaps of progress in modern NLP, the conquest of NLI has proven challenging to this day.
Neural systems show a tendency to abuse spurious data patterns over actually performing the (often complicated) reasoning required to solve the problem, resulting in limited generalization capacity across datasets.
For our dataset, we follow \citet{GLUE,SGLUE} in establishing a hierarchy of rudimentary but descriptive linguistic tags that encompass an array of phenomena that can influence the direction of inference.
For a glimpse at the full hierarchy of tags used, refer to Table~\ref{tab:nli_jaccard}.
These tags are intended to find use outside the model's input/output pipeline, providing a guide for categorizing results and drawing finer-grained quantitative evaluations.
Where our dataset diverges from established practices is in providing an explicit account of inference-level ambiguities not only through the tagging but also through the labeling scheme.
Rather than annotating each example pair with any \textit{one} inference label, we instead specify \textit{all} possible labels that may hold.
To do so, we implicitly consider the product space of all possible readings of both premise and hypothesis, and construct the label set arising out of all pairwise interactions; Figure~\ref{fig:multi_label} shows two concrete examples under different settings.

To create the collection of samples that make up the dataset, we follow a three stage process.
At the first stage, each author independently wrote a number of sentence pairs together with a suggested set of tags and labels.
Afterwards, each author was given a collection of sentence pairs from other authors with the tags and labels hidden, and was tasked with assigning the tags and labels they deemed most appropriate.
This way, we end up with four unique tag and label sets for each pair.
Finally, we perform an aggregation of the proposed annotations and jointly go through any and all examples that contain at least one tag or label that does not reach a majority (i.e. counts less than three votes).
We resolve disagreements by adding or removing annotations, thus ensuring internal consistency within the dataset.
At the end of the process, we end up with 1\,049 samples, of which 110 contain more than a single label.
The dataset as a whole contains 454 \texttt{Neutral}, 414 \texttt{Entailment} and 292 \texttt{Contradiction} assignments.

In parallel to the above, we re-annotate the Greek version of FraCaS~\cite{amanaki-etal-2022-fine} according to our format specifications, skipping directly to the third stage of the pipeline described earlier.
The derived dataset contains an additional 713 examples, revealing 30 of them as multi-labeled, with a label distribution of 264 \texttt{Neutral}, 345 \texttt{Entailment} and 134 \texttt{Contradiction}.
We serve the two datasets independently, but as a single resource.

\begin{figure*}
    \renewcommand{\arraystretch}{0.9}
    \begin{tabularx}{0.995\textwidth}{@{}l@{\quad}l@{}}
        Premise                     & \textgreek{Ο Κυριάκος φίλησε την Αντιγόνη.}\\
                                    & ~\textit{Kyriakos kissed Antigone.} \\
        Hypothesis                  & \textgreek{Ο Κυριάκος και η Αντιγόνη φιλήθηκαν.}\\
                                    & ~\textit{Kyriakos and Antigone kissed [each other].}\\
        Labels                      & \texttt{Entailment}, \texttt{Unknown}\\
        Tags                        & Lexical Entailment:Symmetry/Collectivity\\
        \addlinespace
        Premise                     & \textgreek{Ο Γιώργος είπε στη Μαρία ότι ξέρει να παίζει κιθάρα.}\\
                                    & ~\textit{Giorgos told Maria that [he/she] knows how to play the guitar.} \\
        Hypothesis                  & \textgreek{Η Μαρία ξέρει να παίζει κιθάρα.}\\
                                    & ~\textit{Maria knows how to play the guitar.}\\
        Labels                      & \texttt{Entailment}, \texttt{Unknown}\\
        Tags                        & Lexical Entailment:Factivity:Factive, Predicate-Argument Structure:Anaphora/Coreference\\
    \end{tabularx}
    \caption{NLI examples 761 and 879, showcasing multiple inferences. In the first example, \textgreek{φιλώ} [/fil\'o/] (\textit{to kiss}) can be a unidirectional or a reciprocal action (i.e., \textit{to give a kiss to} vs. \textit{to exchange kisses with}). In the second example, pro-drop allows for two possible readings, where either Giorgos or Maria can be the subject of the embedded clause.}
    \label{fig:multi_label}
\end{figure*}

\subsection{Repurposing the Lexicon}
Transitioning to our next objective, a resource targeting lexical semantics, we immediately run into a roadblock.
The construction of a sufficiently large dataset centered on the \textit{word} requires a prohibitive investment of time and effort.
Facing the very same challenge, contemporary contributions have established the practice of turning to either machine translation or crowd-sourced labor, with hired workers being overlooked by applied practitioners (at best, if at all).
Albeit pragmatic, this approach compromises the quality of the generated resources, dismissing domain expertise in the pursuit of improved cost efficiency (a prerequisite, in turn, for quantity).
As an alternative, we redirect our focus towards a frequently-overlooked traditional resource: the \textit{lexicon}.
Reputable lexica offer a rare mixture of linguistic rigor and extensive coverage virtually for free, making them a prime candidate for adaptation and repurposing into modern applications.
In what follows, we showcase how this insight can be put into practice, enacting a sensible and effective way forward for under-resourced languages.

We begin by procuring a copy of the Dictionary of Standard Modern Greek~\cite{triantafyllides1998dictionary}.%
    \footnote{Hosted online at \url{www.greek-language.gr/greekLang/modern_greek/tools/lexica/triantafyllides}.}
The dictionary is provided in the form of a minimally structured SQL database, associating each lemma with its lexical entry, a raw text field containing a periphrastic definition and a few usage examples for each of its senses.
Unfortunately, senses and examples are not structurally differentiated by the database, but are rather presented in the same field, further intertwined with supplementary details such as usage conditions, morphological information, etc.
Instead, the database relies on a combination of formatting strategies, including enumeration and styling, to differentiate between definitions and examples.
However, these strategies are not consistently applied across the lexicon.
To make matters worse, definitions and examples are often woven together (that is, they materialize as non-contiguous strings), and can at times follow ad-hoc hierarchical arrangements.
Consequently, even though the textual content effectively conveys information visually, parsing this content with traditional methods proves nigh impossible. 
As a workaround, and considering that parsing unstructured data is a staple task for large language models, we employ ChatGPT~\cite{brown2020language} for the problem at hand.

Our pipeline is as follows.
We first utilize the existing database fields to filter the lexical entries that seem to contain at least one example.
This results in a collection of 28\,831 unique lemmata, each mapped to its lexical entry. 
We randomly sample 100 of them, which we then manually convert into a succinct and minimally structured JSON format, specifying (i) the \textit{lemma} and (ii) a list of \textit{senses}, each sense structured as a \textit{definition} and a list of \textit{examples}.
We put extra effort into disentangling hierarchical senses, repeating the elided parts of non-contiguous definitions and examples and removing enumeration identifiers.
The yield of this process then serves as the training set for a quick one-shot tuning of ChatGPT%
\footnote{We use model \texttt{gpt-3.5-turbo} via the fine-tuning API.}, the input being the raw text (stripped of HTML tags for token economy) and the target being the structured JSON representation.
We pass all remaining entries through the trained model.
From the model output, we filter out senses that contain no examples and entries that contain less than two senses, and end up with 16\,079 examples spread over 7\,677 senses and 2\,512 entries.
Finally, we manually check each and every example and entry, throwing away the occasional parsing error, homogenizing the presentation and fixing the JSON formatting as needed.
The result is 14\,416 examples spread over 6\,896 senses and 2\,326 entries, from which we derive the three evaluation tasks described in the subsections to follow.

\paragraph{The Role of ChatGPT} 
Our decision to incorporate a large language model model into our data preparation process does not entail any of the epistemological risks commonly associated with generative models and/or data augmentation.
In our use case, the model does not need a deep understanding of the Greek language, the expertise of a trained linguist, or the creativity required of a human annotator, as it's neither generating new examples nor annotating existing ones per se.
Rather, it suffices for it to recognize the inconsistent yet intuitive hierarchical enumeration patterns present in the data, and to convert them into recurring structures with consistent formatting.
Large language models' attested proficiency in this scenario align them perfectly with our needs, allowing us to utilize the authoritative resource of the lexicon while minimizing tedious human labor and cost expenditure.
Indeed, our inspection of the model's output shows a generally high-quality translation, strictly faithful to the original input, with only a few minor occasional inconsistencies%
\footnote{The model is sometimes overeager, extending the output specification with additional fields, in what seems like an attempt to capture all the information provided in the raw input.}.

\subsubsection{Words-in-Context}
The first task is essentially a replica of the Words-in-Context (WiC) part of SuperGlue.
It is formulated as a binary classification problem, where the system is presented with two sentences containing the same (potentially polysemous) word, and is tasked with telling whether the two occurrences correspond to the same meaning or not.
In order to successfully resolve the task, the system needs a dynamic embedding strategy, capable of disambiguating words depending on their surrounding context.
As such, it serves as a primitive test suite for the lexical semantic capacities of bidirectional transformers.

Obtaining the task from our dataset is trivial; it suffices to consider the sum of the product space of examples for each lexical entry (with the diagonals removed), zipped with a boolean sign indicating whether the two examples stem from the same sense.
Doing so yields 117\,662 data points (i.e., one order of magnitude larger than the corresponding fragment of SuperGlue), with a label ratio of 1 positive to about 6 negative.

\subsubsection{Sense Selection}
The above formulation is straightforward, and directly compatible with the standard sequence classification pipeline commonly employed by NLP architectures.
As such, it makes for an accessible entry point for evaluation.
However, it represents a dramatic simplification of the disambiguation problem, requiring two usages in juxtaposition and providing little information on \textit{what} the sense of each usage is.
Our source dataset allows us to do better.
Given that we have periphrastic definitions for all%
\footnote{Excluding the ones removed by the filtering process.}
the possible meanings of each word, we can reframe the task as sense selection.
Given a word, the set of its possible meanings and a usage context, we can prompt a model to predict the meaning most likely employed in the given context.
Using periphrastic definitions as a proxy for meaning induces a better informed and more realistic evaluation task, requiring and benefiting from high-quality contextual representations both at the lexical and the sentential level (since the word under scrutiny will now need to be contrasted to the full set of ``meanings'').
It is also more faithful to the source dataset, since the count of data points is now in alignment with the number of distinct usage examples (as duplication is no longer necessary).
Each of the 14\,416 points is associated with 3.8 candidate definitions, on average.


\subsubsection{Metaphor Detection}
Our projection of the raw textual entries into structured JSON entries has done away with most fields irrelevant to word disambiguation.
However, we have consciously kept markers of metaphoric usage, and homogenized their presentation.%
\footnote{They are indicated with \textgreek{(μτφ.)} in the periphrastic definition.}
This enables us to filter senses (and by extension, usage examples) that are used metaphorically, providing the means for another kind of task altogether: metaphor detection.
Making the simplifying assumption that metaphor is only present in those examples where the word defined is used in a metaphoric sense, we end up with 1\,017 examples of metaphor (7\% of the total of all examples) concentrated around 571 senses and associated with 499 entries, yielding a heavily imbalanced dataset for metaphor detection.

\section{Experimental Baselines}
To quantitatively evaluate the difficulty of the tasks described in the previous section, and in order to facilitate future research in this direction, we set up some experimental baselines using the current state-of-the-art machinery available for modern Greek.
All our experiments rest on the tried and tested fine-tuning process for BERT-like models~\cite{kenton2019bert}, using Greek BERT as our universal core model~\cite{greekBert}.

\subsection{Natural Language Inference}
Despite our efforts to create a comprehensive evaluation suite for natural language inference, the practical use of our dataset presents several challenges.
First and foremost, its comparatively small size renders it unsuitable for fine-tuning purposes. 
This becomes especially problematic considering the lack of NLI datasets tailored specifically for Greek.
Compounding these challenges is the fact that our dataset utilizes a multi-label setup, which complicates direct cross-dataset evaluations.
To address these challenges, we have chosen to leverage XNLI~\cite{conneau2018xnli}, a cross-lingual dataset for language inference of substantial size; while XNLI was not initially designed for training purposes, it presents a viable solution considering the constraints we face.
We employ an iterative approaching when splitting our dataset, aiming for a 30/70 division and taking care to keep the ratio consistent for each of the linguistic tags used.
We then fine-tune BERT, training on the joined test set of XNLI and the smaller of the two splits, evaluating on the dev set of XNLI, and testing on the larger split.
This setup accounts for domain adaptation, while allowing us to frame the problem as multi-label classification (where the XNLI problems are ``coincidentally'' single-label).

\paragraph{}
Concretely, we independently contextualize the premise and hypothesis sentences, concatenate their \texttt{[CLS]} tokens and project them into three independent logits via an intermediate feed-forward layer of dimensionality 64, gated by the GELU activation function~\cite{hendrycks2016gaussian}.
We train using AdamW~\cite{loshchilov2018decoupled} with a batch size of 32 and a learning rate of 10\textsuperscript{-5}.
Despite heavy regularization (weight decay of 0.1, dropout of 0.33 and early stopping), the model is quick to overfit the training set, with development set performance lagging significantly behind (despite the matching domain).
Since accuracy is no longer a suitable performance metric, owing to the multi-label setup we have adopted, we report per-class precision, recall and F1 scores over the test set instead, averaged over three repetitions.
The results, presented in Table~\ref{tab:nli_f1}, are largely underwhelming, indicative of the difficulty of the dataset and confirming the inadequacy of (the Greek fragment of) XNLI as a training and evaluation resource -- a fact also noted by \citet{evdaimon2023greekbart} and consistent with the comparatively low scores of~\citet{amanaki-etal-2022-fine}.
To gain a better understanding of the trained model's behavior across different linguistic phenomena, we group samples according to their linguistic tags, and measure the average Jaccard similarity coefficient between predicted and true labels (i.e., the length of the intersection over the length of the union between the two sets).
As Table~\ref{tab:nli_jaccard} suggests, performance is consistently low across the board.
The model seems to especially struggle with recognizing the effect of embedded clauses (regardless of whether they are restrictive or not), focus associating operators, non-intersective adjectives, hypo- and hypernymy, antonymy and negation.

\begin{table}
    \centering
    \begin{tabular}{@{}l@{\qquad}ccc@{}}
        \textbf{Label}
          & \textbf{Prec.} 
          & \textbf{Rec.} 
          & \textbf{F1}\\
        \toprule
         ~Unkn.     & \num{0.32}{4.9} & \num{0.41}{1.0} & \num{0.35}{3.7}\\
         ~Ent.     & \num{0.52}{2.8} & \num{0.46}{2.7} & \num{0.48}{1.1}\\
         ~Contr.   & \num{0.20}{0.7} & \num{0.26}{7.6} & \num{0.23}{0.6}
    \end{tabular}
    \caption{Per-label test metrics for NLI.}
    \label{tab:nli_f1}
\end{table}

\begin{table}
    \centering
    \renewcommand{\arraystretch}{0.8865}
    \begin{tabular}{@{}l@{\qquad\qquad}c@{}}
         \multicolumn{2}{@{}l@{}}{\textbf{Tag}\hfill \textbf{Jaccard Index (ave.)}}\\
         \toprule
         \multicolumn{2}{@{}c@{}}{Logic}\\
         \midrule
         ~Disjunction                   & \num{0.32}{3.2}\\
         ~Conjunction                   & \num{0.41}{1.6}\\
         \multicolumn{1}{@{}l}{~Negation}\\
         ~~Single                        & \num{0.30}{1.6}\\
         ~~Multiple                      & \num{0.46}{5.6}\\
         ~~Negative Concord              & \num{0.32}{0.4}\\
         ~Comparatives                      & \num{0.42}{3.5}\\
         \multicolumn{1}{@{}l}{~Quantification}\\
         ~~Existential                  & \num{0.43}{1.0}\\
         ~~Universal                    & \num{0.36}{1.3}\\
         ~~Non-Standard                 & \num{0.37}{2.8}\\
         ~Temporal                      & \num{0.32}{1.1}\\
         ~Conditionals                  & \num{0.32}{3.2}\\
         \addlinespace
         \multicolumn{2}{@{}c@{}}{Lexical Entailment}\\
         \midrule
         ~Redundancy                    & \num{0.33}{1.1}\\
         \multicolumn{1}{@{}l}{~Factivity}\\
         ~~Factive                      & \num{0.41}{2.2}\\
         ~~Non-Factive                  & \num{0.32}{4.0}\\
        \multicolumn{1}{@{}l}{~Intersectivity}\\
         ~~Intersective                 & \num{0.38}{4.2}\\
         ~~Non-Intersective             & \num{0.29}{7.4}\\
         \multicolumn{1}{@{}l}{~Restrictivity}\\
         ~~Restrictive                  & \num{0.28}{2.9}\\
         ~~Non-Restrictive              & \num{0.27}{4.0}\\
         \multicolumn{1}{@{}l}{~Lexical Semantics}\\
         ~~Synonymy                     & \num{0.46}{2.9}\\
         ~~Hyponymy                     & \num{0.47}{1.8}\\
         ~~Hypernymy                    & \num{0.29}{5.6}\\
         ~~Antonymy                     & \num{0.30}{3.2}\\
         ~~Meronymy                     & \num{0.50}{2.5}\\
         ~Morph. Modification           & \num{0.33}{1.8}\\
         ~FAO                           & \num{0.28}{1.3}\\
         ~Symmetry/Collectivity         & \num{0.44}{4.1}\\
         \addlinespace
         \multicolumn{2}{@{}c@{}}{Predicate-Argument Structure}\\
         \midrule
         ~Alternations                  & \num{0.38}{2.0}\\
         ~Ambiguity                     & \num{0.40}{2.9}\\
         ~Anaphora/Coreference          & \num{0.39}{0.1}\\
         ~Ellipsis                      & \num{0.44}{1.7}\\
         ~Core Arguments                & \num{0.55}{5.0}\\
         \addlinespace
         Common Sense/Knowledge         & \num{0.36}{0.3}\\
         \midrule
    \end{tabular}
    \caption{Per-tag test metrics for NLI. The tag hierarchy follows along \citet{SGLUE}, with few divergences. 
    For Logic, we replace Double Negation with Multiple Negations and differentiate it from Negative Concord. We add a tag for Non-Standard Quantification, and drop the Numeral/Interval tag.
    For Lexical Entailment, we substitute Morphological Negation with the (more general) Morphological Modification. We subcategorize Lexical Semantics, specifying left-to-right or premise-to-hypothesis (directional) lexical relations.
    Finally, we merge Common Sense and World Knowledge into a single meta-tag.
    }
    \label{tab:nli_jaccard}
\end{table}

\subsection{Sense Disambiguation}
For both variants of the sense disambiguation task, we split the dataset's examples into three subsets: a 60\% training set, a 20\% development set, and a 20\% test set. 
Additionally, we designate 10\% of the total lexical entries as test-only, and move the associated examples from the training set to the test set. 
This will allow us to evaluate the model's performance separately on in- and out-of-vocabulary examples (IV and OOV, respectively), i.e. involving words that have or have not been encountered during training.

To find the relevant word within each example, we lemmatize examples using SpaCy~\cite[model \texttt{el\_core\_news\_sm}]{Honnibal_spaCy_Industrial-strength_Natural_2020} and identify the element within each sequence that corresponds to the source entry's lemma, falling back to the element with the minimal edit distance if no absolute match can be found.
Following tokenization, this permits us to create a boolean mask for each example, selecting only these tokens that are associated with the word/lemma of interest.

\paragraph{Words-in-Context}
For the WiC variant, we gather minibatches consisting of all examples that belong to the same lexical entry.
We contextualize examples independently, and extract the representations of the words of interest by mean pooling the last layer representations of the tokens selected by each example's mask.
We then compute pairwise similarity scores between pairs in the cartesian product of examples by applying the dot-product operator on the extracted representations, scaling the results by the inverse of the square root of the model's dimensionality.
These similarity scores serve as logits for binary cross entropy training, predicting whether the two occurrences of the word share the same sense between the two examples.

\begin{table*}
    \centering
    \begin{tabular}{@{}l@{\qquad\quad}cc@{\qquad\quad}ccc@{}}
    	                   & \multicolumn{2}{@{}c}{\textbf{Sense Selection}}         & \multicolumn{3}{c@{}}{\textbf{Words-in-Context}}\\
        \textbf{Subset}   & \# examples  & accuracy               & \# pairs        & accuracy\textsuperscript{1}     & accuracy\textsuperscript{2}\\
        \toprule
        ~IV               & 2\,494       & \num{0.63}{0.20}       & 8\,274          & \num{0.50}{0.41} & \num{0.51}{1.7}\\
        ~OOV              & 1\,289       & \num{0.64}{0.41}       & 9\,954          & \num{0.48}{1.77} & \num{0.54}{0.2}\\
        \midrule
        ~Total            & 3\,784       & \num{0.63}{0.29}       & 18\,678         & \num{0.49}{1.09} & \num{0.53}{0.86}\\
        \addlinespace
        \multicolumn{6}{@{}l@{}}{\smaller{\textsuperscript{1}\,In-domain evaluation of the words-in-context model.}}\\
        \multicolumn{6}{@{}l@{}}{\smaller{\textsuperscript{2}\,Transfer evaluation of the sense selection model.}}
    \end{tabular}
    \caption{Test set sizes and performance metrics for the two sense disambiguation tasks.}
    \label{tab:sense_selection}
\end{table*}

\paragraph{Sense Selection}
For the sense selection variant, we create batches by (i) sampling over training examples and (ii) constructing the set union of all related (candidate) definitions, together with a binary boolean relation specifying whether an example and a definition belong to the same entry.
We then independently contextualize all examples and definitions, extracting contextual word representations for each example as before, and taking each definition's \texttt{[CLS]} token representation as a proxy for the sense's meaning.
We compare each word (in the context of a single example) to each meaning using the same scaled dot-product mechanism as before, masking out invalid pairs according to the example-to-definition relation mentioned earlier.
We finally obtain softmax scores for each example yielding a probability distribution over candidate meanings, which serves as the model outputs for standard negative log-likelihood training.

\paragraph{}
We train on either task using AdamW with a learning rate of 10\textsuperscript{-5}, a weight decay of 10\textsuperscript{-2} and a 25\% dropout applied at the dot-product indices, and perform model selection on the basis of development set accuracy; once more, development and training set performances quickly diverge after a few epochs. 
At this point, we note that both tasks use the same notion of sense agreement and both our models approximate it by means of the same vector operation; their difference lies in the fact that one compares a word occurrence to a word occurrence (or: an example to an example), whereas the other compares a word occurrence to a set of ``meanings'' (or: an example to all candidate definitions)~\cite{hauer-kondrak-2022-wic}.
Intuitively, it would make sense that a model that has acquired the sense selection task should be able to perform adequately on the WiC task without further training; indeed, if two word occurrences select the same meaning (i.e., maximize their similarity to the same vector), they must also be similar to one another.
To test this hypothesis, we simply apply the model obtained by fine-tuning on the sense selection task, except now recasting the test set in the form of the WiC task.

We report repetition-averaged aggregates in Table~\ref{tab:sense_selection}.
Performance is not astonishing, but remains well above the random baselines for both tasks (25\% for sense selection and 16.7\% for WiC), indicating that the core model has some capacity for learning and generalization.
Sense selection may initially appear as the more challenging of the two tasks, seeing as it involves selecting one target out of multiple options.
Nonetheless, the model achieves a consistently higher absolute accuracy there; evidently, comparing one example to a fixed set of senses is easier than comparing two ad-hoc usage examples.
To our surprise, we find that the task transfer setup works straight out of the box, to the point where the transfer model in fact outperforms the in-domain model without as much as recalibrating the sigmoid classification threshold.
One might hypothesize that this is due to the model memoizing a fixed set of senses and their representations.
However, this is not entirely the case: interestingly, accuracy now improves instead of declining in the OOV fragment of the test set.
We interpret this as evidencing that the sense selection formulation produces a higher quality error signal, which induces a better informed disambiguation prior during fine-tuning, allowing the (more rudimentary) WiC task to be captured without additional effort.

\subsection{Metaphor Detection}
The last task, metaphor detection, is also the simplest one, being essentially a case of sequence classification.
We start by filtering all entries that have at least one metaphoric sense, so as to alleviate the severe class imbalance of the full dataset.
From the 499 filtered entries, we reserve 5\% for use as an OOV test set.
We extract all examples from all entries, and assign to each example a boolean label, indicating whether the sense the example is associated with is metaphoric or not.
This produces 3\,015 examples (2\,856 IV and 159 OOV), with a class distribution of about 1 positive to 2 negative.
We proceed with training using once more a 60/20/20 split on the IV set.

We attach a feedforward classifier to the contextualized \texttt{[CLS]} token and train using binary cross entropy, optimizing with the same hyper-parameter setup as before.
Our results, presented in Table~\ref{tab:metaphor}, showcase a good ability to recognize metaphoric senses in the words trained on, and a decent generalization potential to unseen words.
Unlike prior experiments, we detect a high variability in the results between repetitions; one model instance has a moderate performance that does not differ between the two subsets of the test set, whereas another achieves a near-perfect score on the IV subset while being barely above the random baseline in the OOV subset.

\begin{table}
    \centering
    \begin{tabular}{@{}l@{\quad}cc@{}}
         \textbf{Subset} & \# Examples & \textbf{Accuracy}\\
         \toprule
         ~IV    & 572 & \num{0.84}{6.29}\\
         ~OOV   & 159 & \num{0.71}{2.94}\\
         \midrule
         ~Total & 731 & \num{0.82}{4.29}
    \end{tabular}
    \caption{Test set performance on the metaphor detection task.}
    \label{tab:metaphor}
\end{table}

\section{Related Work}
NLI is widely considered one of the core problems towards natural language understanding, with a plethora of evaluation suites~\cite{SNLI,conneau2018xnli,GLUE,SGLUE,ANLI} which continue to pose significant challenge for current state-of-the-art models~\cite[\textit{inter alia}]{glockner-etal-2018-breaking,talman-chatzikyriakidis-2019-testing,belinkov-etal-2019-dont,mccoy-etal-2019-right,richardson2020probing}.
Like GLUE and SuperGlue, our inference examples come packed with linguistic tags to facilitate diagnostic analysis.
Unlike other datasets, our examples may specify more than one inference label, accounting for all possible sentence readings.
At the time of writing, other than a fragment of XNLI (produced by automatic translation), the only NLI dataset for Greek we are aware of is by \citet{amanaki-etal-2022-fine} (which we adapt here to our format).

Sense repositories, i.e., mappings between words and sets of meanings are often framed as dictionary-like structures~\cite{fellbaum1998wordnet,navigli2012babelnet}.
Our dataset stands out in providing both a definition and a collection of examples for each sense, allowing the incorporation of either or both into various possible tasks and model pipelines; we show three concrete examples of how this can be accomplished.
The tasks obtained, namely words-in-context, sense selection and metaphor detection, are of prime importance for the experimental validation of the lexical semantic capacities of language processing systems~\cite{ma2021,zhang-liu,choi-etal-2021-melbert,sengupta2022,luo-etal-2023-together}.
To the best of our knowledge, this is the first dataset of its kind, and among the first lexical resources for Greek in general.

\section{Conclusions and Future Work}
Our vision is that of an open-source, community-owned, dynamically adapted, gold-standard suite that enables the linguistically conscious evaluation of the capacities of Greek language models.
We have presented four novel tasks and corresponding baselines towards that goal.
While our results aren't directly comparable to existing benchmarks, they do highlight the significant challenge our tasks present. 
This underscores the urgency for accelerated progress within the Greek NLP ecosystem to stay aligned with contemporary mainstream research. 

Pending community feedback, we hope to enrich the existing datasets by scaling them up, correcting possible artifacts and extending the language domain with regional and dialectal variations.
Possible tasks that we would like the project to eventually incorporate include gender bias detection, paraphrase identification, and natural language inference with explanations, among others.
We are curious to continue experimenting with ways to utilize traditional resources, and exploring their potential as dataset generators for under-resourced languages in conjunction with large language models.

\newpage

\section*{Limitations}
The NLI dataset's limited size renders it inadequate as a comprehensive resource for training and evaluating NLI systems from scratch.
Furthermore, the examples were crafted by the authors of this paper, who belong to a distinct demographic, unavoidably introducing our own cultural, sociopolitical, and linguistic biases.
The focus is exclusively on standard modern Greek, omitting examples of regional or dialectal language use.
Finally, while the tag set employed may provide valuable information, it offers only a coarse and incomplete summary of the full range of linguistic phenomena observed in the wild.

The lexical dataset, conversely, is not indicative of our opinions as authors; the source dictionary may contain language use that is outmoded or socially exclusive.
The dataset structure is sufficient for us to extract the three tasks we have presented, but might prove lacking for more complex tasks (like tasks requiring hierarchical or clustered sense arrangements, for instance).
Despite efforts to ensure semantic accuracy in every entry, sense, and example, occasional mistakes may have gone unnoticed. 
Users should approach the resource critically, keeping this in mind.

Regarding our baselines, we have experimented with only a single model.
While we acknowledge this might entangle the effects of dataset difficulty and model robustness, we justify ourselves in refraining from experimenting with more models, since this is neither the prime concern of this paper, nor a practice that we necessarily agree with.

\section*{Acknowledgements}
We would like to acknowledge the Centre for the Greek Language for allowing us access to the digitized version of the dictionary.
The first author would like to thank Savvas Papadopoulos for sharing his technical expertise on using ChatGPT effectively.
The project has benefited from a grant from the Special Account for Research Funding of the Technical University of Crete (grant number: 11218).

\bibliography{custom}

\begin{thebibliography}{38}
\expandafter\ifx\csname natexlab\endcsname\relax\def\natexlab#1{#1}\fi

\bibitem[{Amanaki et~al.(2022)Amanaki, Bernardy, Chatzikyriakidis, Cooper,
  Dobnik, Karimi, Ek, Giannikouri, Katsouli, Kolokousis, Mamatzaki, Papadakis,
  Petrova, Psaltaki, Soupiona, Skoulataki, and
  Stefanidou}]{amanaki-etal-2022-fine}
Eirini Amanaki, Jean-Philippe Bernardy, Stergios Chatzikyriakidis, Robin
  Cooper, Simon Dobnik, Aram Karimi, Adam Ek, Eirini~Chrysovalantou
  Giannikouri, Vasiliki Katsouli, Ilias Kolokousis, Eirini~Chrysovalantou
  Mamatzaki, Dimitrios Papadakis, Olga Petrova, Erofili Psaltaki, Charikleia
  Soupiona, Effrosyni Skoulataki, and Christina Stefanidou. 2022.
\newblock \href {https://aclanthology.org/2022.dclrl-1.6} {Fine-grained
  entailment: Resources for {G}reek {NLI} and precise entailment}.
\newblock In \emph{Proceedings of the Workshop on Dataset Creation for
  Lower-Resourced Languages within the 13th Language Resources and Evaluation
  Conference}, pages 44--52, Marseille, France. European Language Resources
  Association.

\bibitem[{Artetxe et~al.(2020)Artetxe, Labaka, and
  Agirre}]{artetxe-etal-2020-translation}
Mikel Artetxe, Gorka Labaka, and Eneko Agirre. 2020.
\newblock \href {https://doi.org/10.18653/v1/2020.emnlp-main.618} {Translation
  artifacts in cross-lingual transfer learning}.
\newblock In \emph{Proceedings of the 2020 Conference on Empirical Methods in
  Natural Language Processing (EMNLP)}, pages 7674--7684, Online. Association
  for Computational Linguistics.

\bibitem[{Belinkov et~al.(2019)Belinkov, Poliak, Shieber, Van~Durme, and
  Rush}]{belinkov-etal-2019-dont}
Yonatan Belinkov, Adam Poliak, Stuart Shieber, Benjamin Van~Durme, and
  Alexander Rush. 2019.
\newblock \href {https://doi.org/10.18653/v1/P19-1084} {Don{'}t take the
  premise for granted: Mitigating artifacts in natural language inference}.
\newblock In \emph{Proceedings of the 57th Annual Meeting of the Association
  for Computational Linguistics}, pages 877--891, Florence, Italy. Association
  for Computational Linguistics.

\bibitem[{Bowman et~al.(2015)Bowman, Angeli, Potts, and Manning}]{SNLI}
Samuel~R Bowman, Gabor Angeli, Christopher Potts, and Christopher~D Manning.
  2015.
\newblock A large annotated corpus for learning natural language inference.
\newblock \emph{arXiv preprint arXiv:1508.05326}.

\bibitem[{Brown et~al.(2020)Brown, Mann, Ryder, Subbiah, Kaplan, Dhariwal,
  Neelakantan, Shyam, Sastry, Askell et~al.}]{brown2020language}
Tom Brown, Benjamin Mann, Nick Ryder, Melanie Subbiah, Jared~D Kaplan, Prafulla
  Dhariwal, Arvind Neelakantan, Pranav Shyam, Girish Sastry, Amanda Askell,
  et~al. 2020.
\newblock Language models are few-shot learners.
\newblock \emph{Advances in neural information processing systems},
  33:1877--1901.

\bibitem[{Choi et~al.(2021)Choi, Lee, Choi, Park, Lee, Lee, and
  Lee}]{choi-etal-2021-melbert}
Minjin Choi, Sunkyung Lee, Eunseong Choi, Heesoo Park, Junhyuk Lee, Dongwon
  Lee, and Jongwuk Lee. 2021.
\newblock \href {https://doi.org/10.18653/v1/2021.naacl-main.141} {{M}el{BERT}:
  Metaphor detection via contextualized late interaction using metaphorical
  identification theories}.
\newblock In \emph{Proceedings of the 2021 Conference of the North American
  Chapter of the Association for Computational Linguistics: Human Language
  Technologies}, pages 1763--1773, Online. Association for Computational
  Linguistics.

\bibitem[{Conneau et~al.(2018)Conneau, Rinott, Lample, Williams, Bowman,
  Schwenk, and Stoyanov}]{conneau2018xnli}
Alexis Conneau, Ruty Rinott, Guillaume Lample, Adina Williams, Samuel~R.
  Bowman, Holger Schwenk, and Veselin Stoyanov. 2018.
\newblock {XNLI}: Evaluating cross-lingual sentence representations.
\newblock In \emph{Proceedings of the 2018 Conference on Empirical Methods in
  Natural Language Processing}. Association for Computational Linguistics.

\bibitem[{Espa{\~n}a-Bonet and
  Barr{\'o}n-Cede{\~n}o(2022)}]{espana-bonet-barron-cedeno-2022-undesired}
Cristina Espa{\~n}a-Bonet and Alberto Barr{\'o}n-Cede{\~n}o. 2022.
\newblock \href {https://doi.org/10.18653/v1/2022.emnlp-main.133} {The
  (undesired) attenuation of human biases by multilinguality}.
\newblock In \emph{Proceedings of the 2022 Conference on Empirical Methods in
  Natural Language Processing}, pages 2056--2077, Abu Dhabi, United Arab
  Emirates. Association for Computational Linguistics.

\bibitem[{Evdaimon et~al.(2023)Evdaimon, Abdine, Xypolopoulos, Outsios,
  Vazirgiannis, and Stamou}]{evdaimon2023greekbart}
Iakovos Evdaimon, Hadi Abdine, Christos Xypolopoulos, Stamatis Outsios,
  Michalis Vazirgiannis, and Giorgos Stamou. 2023.
\newblock Greek{BART}: The first pretrained {G}reek sequence-to-sequence model.
\newblock \emph{arXiv preprint arXiv:2304.00869}.

\bibitem[{Fellbaum(1998)}]{fellbaum1998wordnet}
Christiane Fellbaum. 1998.
\newblock \emph{WordNet: An electronic lexical database}.
\newblock MIT press.

\bibitem[{Glockner et~al.(2018)Glockner, Shwartz, and
  Goldberg}]{glockner-etal-2018-breaking}
Max Glockner, Vered Shwartz, and Yoav Goldberg. 2018.
\newblock \href {https://doi.org/10.18653/v1/P18-2103} {Breaking {NLI} systems
  with sentences that require simple lexical inferences}.
\newblock In \emph{Proceedings of the 56th Annual Meeting of the Association
  for Computational Linguistics (Volume 2: Short Papers)}, pages 650--655,
  Melbourne, Australia. Association for Computational Linguistics.

\bibitem[{Hauer and Kondrak(2022)}]{hauer-kondrak-2022-wic}
Bradley Hauer and Grzegorz Kondrak. 2022.
\newblock \href {https://doi.org/10.18653/v1/2022.naacl-main.178} {{W}i{C} =
  {TSV} = {WSD}: On the equivalence of three semantic tasks}.
\newblock In \emph{Proceedings of the 2022 Conference of the North American
  Chapter of the Association for Computational Linguistics: Human Language
  Technologies}, pages 2478--2486, Seattle, United States. Association for
  Computational Linguistics.

\bibitem[{Havaldar et~al.(2023)Havaldar, Singhal, Rai, Liu, Guntuku, and
  Ungar}]{havaldar-etal-2023-multilingual}
Shreya Havaldar, Bhumika Singhal, Sunny Rai, Langchen Liu, Sharath~Chandra
  Guntuku, and Lyle Ungar. 2023.
\newblock \href {https://doi.org/10.18653/v1/2023.wassa-1.19} {Multilingual
  language models are not multicultural: A case study in emotion}.
\newblock In \emph{Proceedings of the 13th Workshop on Computational Approaches
  to Subjectivity, Sentiment, {\&} Social Media Analysis}, pages 202--214,
  Toronto, Canada. Association for Computational Linguistics.

\bibitem[{Hendrycks and Gimpel(2016)}]{hendrycks2016gaussian}
Dan Hendrycks and Kevin Gimpel. 2016.
\newblock Gaussian error linear units (gelus).
\newblock \emph{arXiv preprint arXiv:1606.08415}.

\bibitem[{Honnibal et~al.(2020)Honnibal, Montani, Van~Landeghem, and
  Boyd}]{Honnibal_spaCy_Industrial-strength_Natural_2020}
Matthew Honnibal, Ines Montani, Sofie Van~Landeghem, and Adriane Boyd. 2020.
\newblock \href {https://doi.org/10.5281/zenodo.1212303} {{spaCy:
  Industrial-strength Natural Language Processing in Python}}.

\bibitem[{Joshi et~al.(2020)Joshi, Santy, Budhiraja, Bali, and
  Choudhury}]{joshi-etal-2020-state}
Pratik Joshi, Sebastin Santy, Amar Budhiraja, Kalika Bali, and Monojit
  Choudhury. 2020.
\newblock \href {https://doi.org/10.18653/v1/2020.acl-main.560} {The state and
  fate of linguistic diversity and inclusion in the {NLP} world}.
\newblock In \emph{Proceedings of the 58th Annual Meeting of the Association
  for Computational Linguistics}, pages 6282--6293, Online. Association for
  Computational Linguistics.

\bibitem[{Kenton and Toutanova(2019)}]{kenton2019bert}
Jacob Devlin Ming-Wei~Chang Kenton and Lee~Kristina Toutanova. 2019.
\newblock {BERT}: Pre-training of deep bidirectional transformers for language
  understanding.
\newblock In \emph{Proceedings of NAACL-HLT}, pages 4171--4186.

\bibitem[{Koutsikakis et~al.(2020)Koutsikakis, Chalkidis, Malakasiotis, and
  Androutsopoulos}]{greekBert}
John Koutsikakis, Ilias Chalkidis, Prodromos Malakasiotis, and Ion
  Androutsopoulos. 2020.
\newblock \href {https://doi.org/10.1145/3411408.3411440} {Greek-{BERT}: The
  {G}reeks visiting sesame street}.
\newblock In \emph{11th Hellenic Conference on Artificial Intelligence}, SETN
  2020, page 110–117, New York, NY, USA. Association for Computing Machinery.

\bibitem[{Kreutzer et~al.(2022)Kreutzer, Caswell, Wang, Wahab, van Esch,
  Ulzii-Orshikh, Tapo, Subramani, Sokolov, Sikasote, Setyawan, Sarin, Samb,
  Sagot, Rivera, Rios, Papadimitriou, Osei, Suarez, Orife, Ogueji, Rubungo,
  Nguyen, Müller, Müller, Muhammad, Muhammad, Mnyakeni, Mirzakhalov,
  Matangira, Leong, Lawson, Kudugunta, Jernite, Jenny, Firat, Dossou, Dlamini,
  de~Silva, Çabuk Ballı, Biderman, Battisti, Baruwa, Bapna, Baljekar, Azime,
  Awokoya, Ataman, Ahia, Ahia, Agrawal, and Adeyemi}]{10.1162/tacl_a_00447}
Julia Kreutzer, Isaac Caswell, Lisa Wang, Ahsan Wahab, Daan van Esch,
  Nasanbayar Ulzii-Orshikh, Allahsera Tapo, Nishant Subramani, Artem Sokolov,
  Claytone Sikasote, Monang Setyawan, Supheakmungkol Sarin, Sokhar Samb,
  Benoît Sagot, Clara Rivera, Annette Rios, Isabel Papadimitriou, Salomey
  Osei, Pedro~Ortiz Suarez, Iroro Orife, Kelechi Ogueji, Andre~Niyongabo
  Rubungo, Toan~Q. Nguyen, Mathias Müller, André Müller, Shamsuddeen~Hassan
  Muhammad, Nanda Muhammad, Ayanda Mnyakeni, Jamshidbek Mirzakhalov,
  Tapiwanashe Matangira, Colin Leong, Nze Lawson, Sneha Kudugunta, Yacine
  Jernite, Mathias Jenny, Orhan Firat, Bonaventure F.~P. Dossou, Sakhile
  Dlamini, Nisansa de~Silva, Sakine Çabuk Ballı, Stella Biderman, Alessia
  Battisti, Ahmed Baruwa, Ankur Bapna, Pallavi Baljekar, Israel~Abebe Azime,
  Ayodele Awokoya, Duygu Ataman, Orevaoghene Ahia, Oghenefego Ahia, Sweta
  Agrawal, and Mofetoluwa Adeyemi. 2022.
\newblock \href {https://doi.org/10.1162/tacl_a_00447} {{Quality at a Glance:
  An Audit of Web-Crawled Multilingual Datasets}}.
\newblock \emph{Transactions of the Association for Computational Linguistics},
  10:50--72.

\bibitem[{Loshchilov and Hutter(2018)}]{loshchilov2018decoupled}
Ilya Loshchilov and Frank Hutter. 2018.
\newblock Decoupled weight decay regularization.
\newblock In \emph{International Conference on Learning Representations}.

\bibitem[{Luo et~al.(2023)Luo, Zhou, and Bollegala}]{luo-etal-2023-together}
Haochen Luo, Yi~Zhou, and Danushka Bollegala. 2023.
\newblock \href {https://doi.org/10.18653/v1/2023.findings-acl.165} {Together
  we make sense{--}learning meta-sense embeddings}.
\newblock In \emph{Findings of the Association for Computational Linguistics:
  ACL 2023}, pages 2638--2651, Toronto, Canada. Association for Computational
  Linguistics.

\bibitem[{Ma et~al.(2021)Ma, Liu, Wang, and Vosoughi}]{ma2021}
Weicheng Ma, Ruibo Liu, Lili Wang, and Soroush Vosoughi. 2021.
\newblock Improvements and extensions on metaphor detection.
\newblock In \emph{Proceedings of the 1st Workshop on Understanding Implicit
  and Underspecified Language}, pages 33--42.

\bibitem[{McCoy et~al.(2019)McCoy, Pavlick, and Linzen}]{mccoy-etal-2019-right}
Tom McCoy, Ellie Pavlick, and Tal Linzen. 2019.
\newblock \href {https://doi.org/10.18653/v1/P19-1334} {Right for the wrong
  reasons: Diagnosing syntactic heuristics in natural language inference}.
\newblock In \emph{Proceedings of the 57th Annual Meeting of the Association
  for Computational Linguistics}, pages 3428--3448, Florence, Italy.
  Association for Computational Linguistics.

\bibitem[{Navigli and Ponzetto(2012)}]{navigli2012babelnet}
Roberto Navigli and Simone~Paolo Ponzetto. 2012.
\newblock Babelnet: The automatic construction, evaluation and application of a
  wide-coverage multilingual semantic network.
\newblock \emph{Artificial intelligence}, 193:217--250.

\bibitem[{Nie et~al.(2020)Nie, Williams, Dinan, Bansal, Weston, and
  Kiela}]{ANLI}
Yixin Nie, Adina Williams, Emily Dinan, Mohit Bansal, Jason Weston, and Douwe
  Kiela. 2020.
\newblock Adversarial nli: A new benchmark for natural language understanding.
\newblock In \emph{Proceedings of the 58th Annual Meeting of the Association
  for Computational Linguistics}, pages 4885--4901.

\bibitem[{Ogueji et~al.(2021)Ogueji, Zhu, and Lin}]{ogueji-etal-2021-small}
Kelechi Ogueji, Yuxin Zhu, and Jimmy Lin. 2021.
\newblock \href {https://doi.org/10.18653/v1/2021.mrl-1.11} {Small data? {N}o
  problem! {E}xploring the viability of pretrained multilingual language models
  for low-resourced languages}.
\newblock In \emph{Proceedings of the 1st Workshop on Multilingual
  Representation Learning}, pages 116--126, Punta Cana, Dominican Republic.
  Association for Computational Linguistics.

\bibitem[{Papadimitriou et~al.(2023)Papadimitriou, Lopez, and
  Jurafsky}]{papadimitriou-etal-2023-multilingual-bert}
Isabel Papadimitriou, Kezia Lopez, and Dan Jurafsky. 2023.
\newblock \href {https://aclanthology.org/2023.sigtyp-1.16} {Multilingual
  {BERT} has an accent: Evaluating {E}nglish influences on fluency in
  multilingual models}.
\newblock In \emph{Proceedings of the 5th Workshop on Research in Computational
  Linguistic Typology and Multilingual NLP}, pages 143--146, Dubrovnik,
  Croatia. Association for Computational Linguistics.

\bibitem[{Pfeiffer et~al.(2021)Pfeiffer, Vuli{\'c}, Gurevych, and
  Ruder}]{pfeiffer-etal-2021-unks}
Jonas Pfeiffer, Ivan Vuli{\'c}, Iryna Gurevych, and Sebastian Ruder. 2021.
\newblock \href {https://doi.org/10.18653/v1/2021.emnlp-main.800} {{UNK}s
  everywhere: {A}dapting multilingual language models to new scripts}.
\newblock In \emph{Proceedings of the 2021 Conference on Empirical Methods in
  Natural Language Processing}, pages 10186--10203, Online and Punta Cana,
  Dominican Republic. Association for Computational Linguistics.

\bibitem[{Richardson et~al.(2020)Richardson, Hu, Moss, and
  Sabharwal}]{richardson2020probing}
Kyle Richardson, Hai Hu, Lawrence Moss, and Ashish Sabharwal. 2020.
\newblock Probing natural language inference models through semantic fragments.
\newblock In \emph{Proceedings of the AAAI Conference on Artificial
  Intelligence}, volume~34, pages 8713--8721.

\bibitem[{Sengupta et~al.(2022)Sengupta, Alshomary, and
  Wachsmuth}]{sengupta2022}
Meghdut Sengupta, Milad Alshomary, and Henning Wachsmuth. 2022.
\newblock Back to the roots: Predicting the source domain of metaphors using
  contrastive learning.
\newblock In \emph{Proceedings of the 3rd Workshop on Figurative Language
  Processing (FLP)}, pages 137--142.

\bibitem[{Talman and
  Chatzikyriakidis(2019)}]{talman-chatzikyriakidis-2019-testing}
Aarne Talman and Stergios Chatzikyriakidis. 2019.
\newblock \href {https://doi.org/10.18653/v1/W19-4810} {Testing the
  generalization power of neural network models across {NLI} benchmarks}.
\newblock In \emph{Proceedings of the 2019 ACL Workshop BlackboxNLP: Analyzing
  and Interpreting Neural Networks for NLP}, pages 85--94, Florence, Italy.
  Association for Computational Linguistics.

\bibitem[{Triantafyllides(1998)}]{triantafyllides1998dictionary}
G~Triantafyllides. 1998.
\newblock Dictionary of standard modern {G}reek.
\newblock \emph{Institute for Modern Greek Studies of the Aristotle University
  of Thessaloniki}.

\bibitem[{Wang et~al.(2019)Wang, Pruksachatkun, Nangia, Singh, Michael, Hill,
  Levy, and Bowman}]{SGLUE}
Alex Wang, Yada Pruksachatkun, Nikita Nangia, Amanpreet Singh, Julian Michael,
  Felix Hill, Omer Levy, and Samuel Bowman. 2019.
\newblock Super{GLUE}: A stickier benchmark for general-purpose language
  understanding systems.
\newblock \emph{Advances in neural information processing systems}, 32.

\bibitem[{Wang et~al.(2018)Wang, Singh, Michael, Hill, Levy, and Bowman}]{GLUE}
Alex Wang, Amanpreet Singh, Julian Michael, Felix Hill, Omer Levy, and Samuel~R
  Bowman. 2018.
\newblock {GLUE}: A multi-task benchmark and analysis platform for natural
  language understanding.
\newblock \emph{arXiv preprint arXiv:1804.07461}.

\bibitem[{Wang and Hershcovich(2023)}]{wang-hershcovich-2023-evaluating}
Zi~Wang and Daniel Hershcovich. 2023.
\newblock \href {https://doi.org/10.18653/v1/2023.acl-long.93} {On evaluating
  multilingual compositional generalization with translated datasets}.
\newblock In \emph{Proceedings of the 61st Annual Meeting of the Association
  for Computational Linguistics (Volume 1: Long Papers)}, pages 1669--1687,
  Toronto, Canada. Association for Computational Linguistics.

\bibitem[{Wu and Dredze(2020)}]{wu-dredze-2020-languages}
Shijie Wu and Mark Dredze. 2020.
\newblock \href {https://doi.org/10.18653/v1/2020.repl4nlp-1.16} {Are all
  languages created equal in multilingual {BERT}?}
\newblock In \emph{Proceedings of the 5th Workshop on Representation Learning
  for NLP}, pages 120--130, Online. Association for Computational Linguistics.

\bibitem[{Yu et~al.(2022)Yu, Chatterjee, Asai, Hu, and
  Choi}]{yu-etal-2022-beyond}
Xinyan Yu, Trina Chatterjee, Akari Asai, Junjie Hu, and Eunsol Choi. 2022.
\newblock \href {https://doi.org/10.18653/v1/2022.findings-emnlp.273} {Beyond
  counting datasets: A survey of multilingual dataset construction and
  necessary resources}.
\newblock In \emph{Findings of the Association for Computational Linguistics:
  EMNLP 2022}, pages 3725--3743, Abu Dhabi, United Arab Emirates. Association
  for Computational Linguistics.

\bibitem[{Zhang and Liu(2023)}]{zhang-liu}
Shenglong Zhang and Ying Liu. 2023.
\newblock \href {https://doi.org/10.18653/v1/2023.findings-acl.96} {Adversarial
  multi-task learning for end-to-end metaphor detection}.
\newblock In \emph{Findings of the Association for Computational Linguistics:
  ACL 2023}, pages 1483--1497, Toronto, Canada. Association for Computational
  Linguistics.

\end{thebibliography}
\bibliographystyle{acl_natbib}

\end{document}